\documentclass[11pt]{article}

\usepackage[preprint]{acl}

\usepackage{times}
\usepackage{latexsym}
\usepackage{amssymb}
\usepackage[T1]{fontenc}

\usepackage[utf8]{inputenc}

\usepackage{microtype}

\usepackage{inconsolata}

\usepackage{xcolor}
\usepackage{graphicx}

\usepackage{amsmath}
\usepackage{multirow}   
\usepackage{booktabs}  
\usepackage{array}
\usepackage{subcaption}
\usepackage{booktabs}
\usepackage{multirow}
\usepackage{caption}
\usepackage{array}
\usepackage{placeins}

\newcolumntype{L}[1]{>{\raggedright\arraybackslash}p{#1}}
\newcolumntype{C}[1]{>{\centering\arraybackslash}p{#1}}

\title{CausalChapter: Improving Long-Video Chaptering with Interventional Dependency Modeling}

\author{Xinran Duan, Guozhang Li, Yaoyao Zhong, Mei Wang\textsuperscript{*}, Lizhi Wang, Hua Huang \\
        School of Artificial Intelligence, Beijing Normal University\\
 Beijing Key Laboratory of Artificial Intelligence for Education\\
 Engineering Research Center of Intelligent Technology and Educational Application, Ministry of Education\\
 \texttt{duanxinran@mail.bnu.edu.cn}, \texttt{\{liguozhang097, zhongyy\}@bnu.edu.cn} \\ \texttt{\{wangmei1, wanglizhi, huahuang\}@bnu.edu.cn}
 }

\begin{document}
\maketitle
\begingroup
\renewcommand\thefootnote{}
\footnotetext{Guozhang Li contributed equally to this work. Corresponding author: Mei Wang.}
\addtocounter{footnote}{-1}
\endgroup
\begin{abstract}
Long-form instructional videos require automatic chaptering to support browsing, navigation, and knowledge access. 
Recent long-context language models can perform chaptering from textualized video inputs, but they remain costly and brittle for content-dense lecture videos with long transcripts, smooth topic transitions, and detailed chapter outputs. 
A scalable segment-then-caption paradigm reduces this cost, but introduces two new challenges: boundary error propagation and fragmented cross-chapter context. 
We propose \textbf{CausalChapter}, an intervention-inspired framework for long-video chaptering that estimates prediction-level influence through lightweight masking and removal interventions. 
For boundary localization, our Local Dependency Shift module detects drops in predictive dependency between adjacent temporal windows; 
for chapter description generation, our Cross-Segment Support Selection module reranks historical contexts according to their support for the current prediction. 
Experiments on long-video chaptering benchmarks show that CausalChapter improves boundary localization, chapter description quality, and cross-chapter coherence.
\href{https://github.com/kong-johnny/CausalChapter}{\textcolor{blue}CausalChapter} 
\end{abstract}

\begin{figure}[t]    
    \centering
    \begin{subfigure}{\columnwidth}
        \centering
        \includegraphics[width=\linewidth]{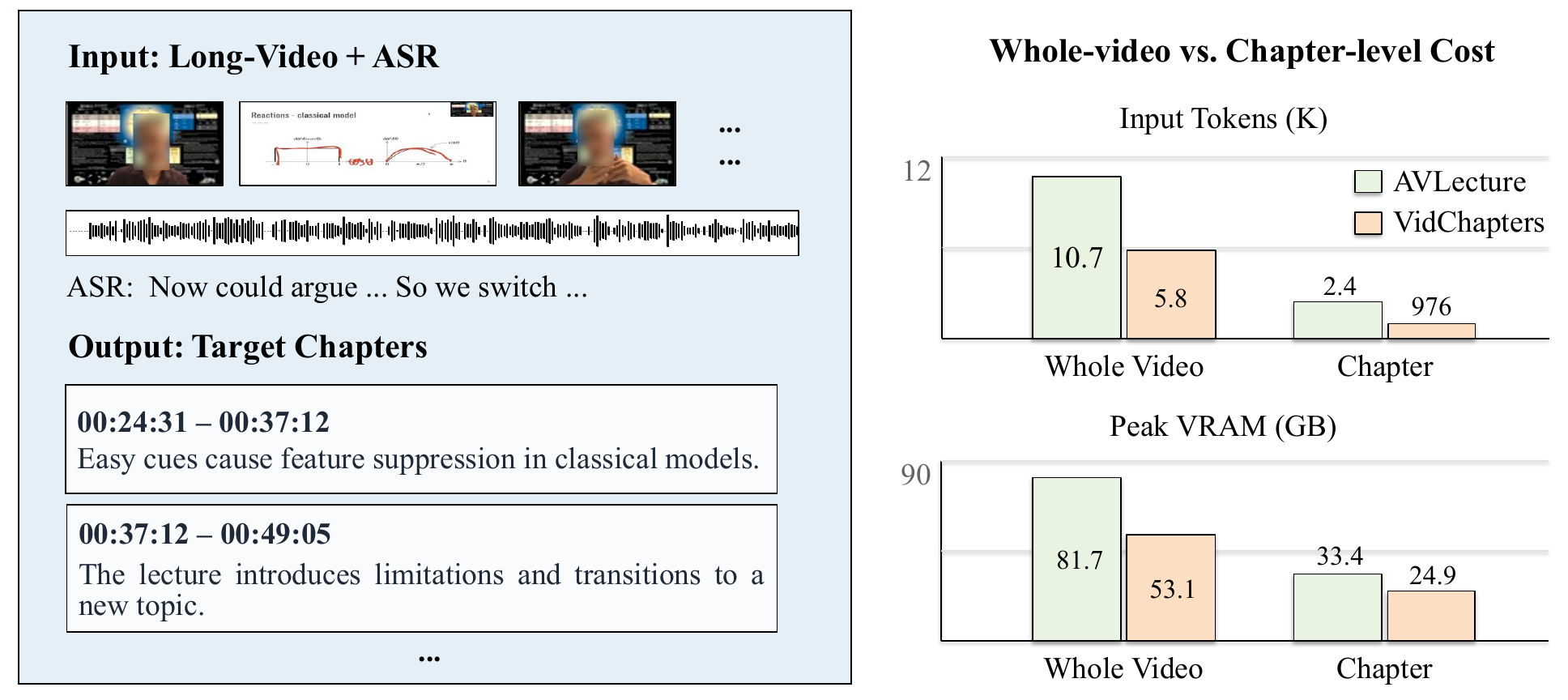}
        \caption{Cost comparison of whole-video and chapter-level inputs.}
    \end{subfigure}    
    \begin{subfigure}{\columnwidth}
        \centering
        \includegraphics[width=\linewidth]{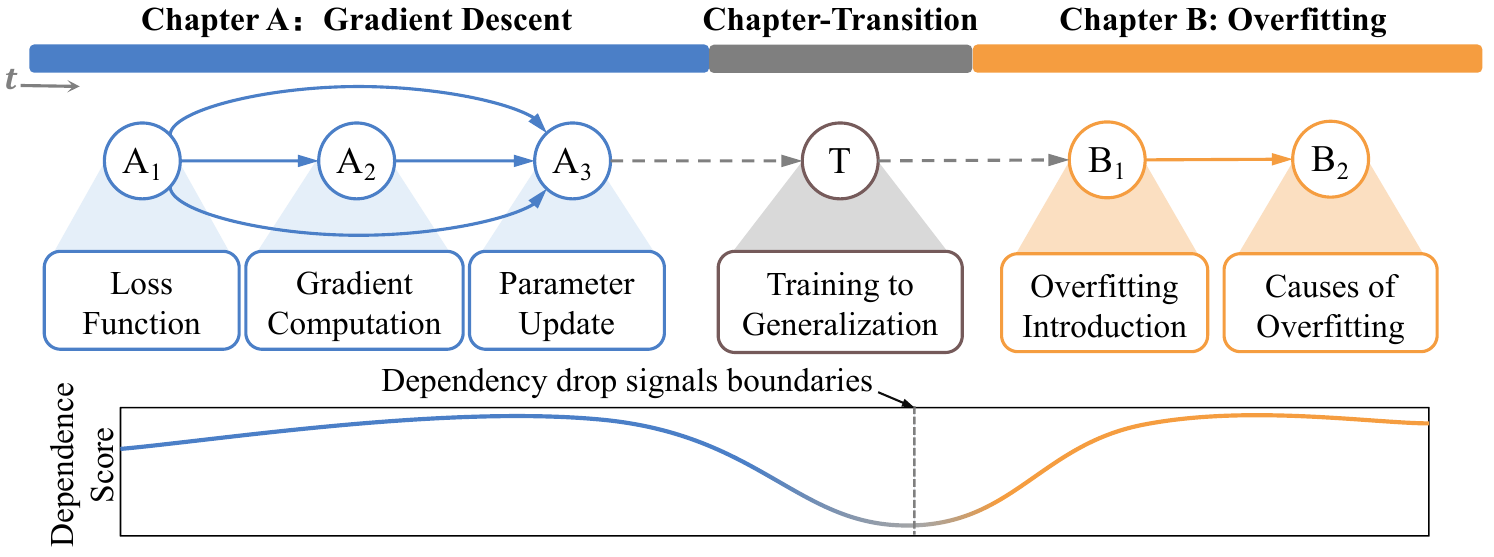}
        \caption{Dependency-shift signals motivate boundary localization.}
    \end{subfigure}    
    \begin{subfigure}{\columnwidth}
         \includegraphics[width=\linewidth]{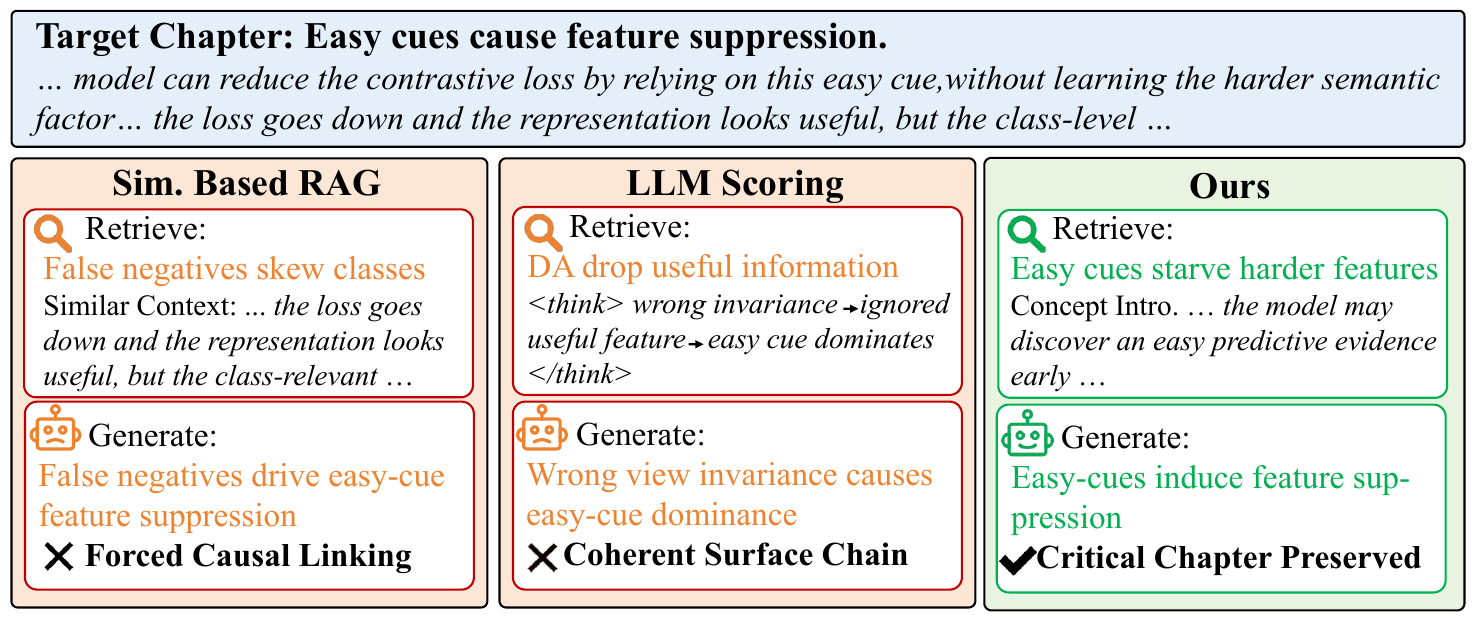}
        \caption{Dependency-aware retrieval preserves predictive cues.}
    \end{subfigure}

    \caption{
Motivation of scalable and dependency-aware long-video chaptering. (a) Content-dense instructional videos require detailed chapter descriptions over long ASR transcripts, making whole-video modeling costly in tokens and memory.
(b) Smooth lecture transitions can preserve local coherence while predictive dependency drops near the true chapter boundary.
(c) Plausible retrieved contexts can induce mechanism drift in chapter generation, whereas prediction-critical context preserves the intended explanation. }
    \label{fig:motivation}
\end{figure}

\section{Introduction}

Long-form videos, such as lectures, meeting recordings, and online tutorials, have become an important source of knowledge and communication. 
To make such videos searchable and navigable, models need to organize them into temporally grounded and semantically coherent units. 
Dense video captioning (DVC) provides a representative localize-and-describe paradigm, where a model detects multiple events in an untrimmed video and generates a description for each event~\citep{krishna2017dense,iashin2020better,wang2021end,yang2023vid2seq}. 
However, DVC mainly focuses on short-term event-level segments, whereas long-form video navigation requires higher-level temporal organization.
Video chaptering addresses this need by partitioning a long video into consecutive chapters and generating navigable titles or descriptions for each chapter~\citep{yang2023vidchapters,ventura2025chapter}.

Recent progress in long-video chaptering has been driven by large-scale datasets and long-context language models.
VidChapters-7M introduces a large-scale benchmark of user-annotated chapters for open-domain videos~\citep{yang2023vidchapters}. 
Chapter-Llama converts videos into timestamped Automatic Speech Recognition (ASR) transcripts and frame captions, and predicts chapter boundaries and free-form titles with a long-context LLM in a single forward pass~\citep{ventura2025chapter}. 
While effective, such holistic long-context modeling faces increasing cost in content-dense instructional videos, where ASR transcripts are long, topic transitions are smooth, and chapter outputs often require detailed descriptions rather than short titles, as illustrated in Fig.~\ref{fig:motivation}(a).
Under practical context budgets, processing the entire textualized video may require truncation, sparse sampling, or compression~\citep{wang2021end,yang2023vid2seq,kim2024you}, which may discard fine-grained evidence needed for boundary localization and chapter description generation.

A scalable alternative is the segment-then-caption paradigm~\citep{islam2024video,zala2023hierarchical}: the model first predicts chapter boundaries to divide a long video into local segments, and then generates a description for each segment. 
This decomposition reduces the input length of each generation step and aligns the generation input with chapter-level outputs. 
However, it also shifts the central challenge from holistic encoding to identifying which local transitions and historical segments truly matter for prediction. 
First, boundary errors may propagate to the generation stage: a shifted boundary can mix adjacent chapter content into the current segment or omit key semantic units. 
Second, independent segment-level generation can break cross-chapter context, which is often crucial in lectures and tutorials where later chapters depend on earlier definitions, assumptions, or examples.

A straightforward solution is to augment each segment with additional context retrieved from nearby or semantically similar segments. 
Existing retrieval- or memory-augmented methods commonly select context based on semantic similarity, temporal proximity, or visual similarity~\citep{kim2024you,kim2025hicm2}. 
However, in instructional long videos, surface relevance does not necessarily imply prediction utility. 
A highly similar segment may simply repeat the current content, while an earlier segment with lower lexical or visual similarity may introduce a definition or logical premise that is crucial for describing the current chapter, as shown in Figure~\ref{fig:motivation}(c). 
This suggests a different criterion for long-video chaptering: a temporal unit should be judged by whether intervening on it changes the model's prediction, rather than by whether it is visually similar, temporally close, or lexically overlapping with the current segment. 
The same criterion can also inform boundary localization. Within a coherent chapter, preceding units usually provide predictive support for subsequent units; near a chapter transition, this dependency may drop even when the transition is visually or lexically smooth, as shown in Figure~\ref{fig:motivation}(b).

Motivated by this observation, we propose \textbf{CausalChapter}, an intervention-inspired framework for scalable long-video chapter generation. 
CausalChapter estimates \emph{intervention-defined predictive dependency}, namely the observable change in model prediction caused by masking, perturbing, or removing input components. 
Our goal is not to recover the real-world causal structure of video content, but to measure the prediction-level influence of semantic units and context segments as a practical signal of predictive support. 
For boundary localization, we introduce \textbf{Local Dependency Shift} (LCDS), which masks semantic units in a preceding temporal window and measures how much the intervention affects the reconstruction of the subsequent window. 
A local drop in this dependency provides an auxiliary signal for detecting smooth chapter transitions. 
For chapter description generation, we introduce \textbf{Cross-Segment Support Selection} (CSSE), which removes candidate context segments and measures their influence on the generated description. 
Segments with high estimated support are selected for second-pass generation, improving the completeness and cross-chapter coherence of the final chapter descriptions.

Our contributions are summarized as follows: (1) We introduce LCDS, an intervention-inspired dependency-drop signal for smooth chapter boundary localization. (2) We introduce CSSE, an intervention-based support estimation mechanism for cross-segment context selection. (3) We show that the proposed interventional mechanism mitigates boundary error propagation and context fragmentation in scalable long-video chaptering.

\section{Related Work} 

\paragraph{Video Chaptering.}
Video chaptering aims to partition a long video into consecutive, non-overlapping, and semantically coherent chapters, while generating titles or summaries for browsing and navigation. 
A related task is dense video captioning (DVC), which localizes and describes multiple events in untrimmed videos~\citep{iashin2020better,iashin2020multi,yang2023vid2seq,wang2021end,kim2024you,liu2025task,wu2025event,xie2025exploring,li2025etc}. 
Recent DVC studies further incorporate memory or retrieval-augmented mechanisms to improve event-level descriptions~\citep{kim2024you,xie2025exploring,wu2025event,liu2025task}.
Although DVC follows a localize-and-describe paradigm, it mainly focuses on short-term event-level segments whose boundaries are often associated with local visual or event changes, making it less suited to long-form videos governed by high-level topic progression.

For long-form video chaptering, VidChapters-7M introduces a large-scale dataset of user-annotated chapters and defines several chaptering tasks, including chapter generation and chapter grounding~\citep{yang2023vidchapters}. 
More recently, Chapter-Llama represents long videos as timestamped ASR transcripts and frame captions, and uses a long-context LLM to jointly predict chapter boundaries and free-form chapter titles in a single forward pass~\citep{ventura2025chapter}. 
These studies demonstrate the effectiveness of textualized video representations and long-context reasoning. 
Different from holistic long-context chaptering methods, we study a scalable segment-then-caption formulation and explicitly address the boundary error propagation and cross-segment context fragmentation introduced by this decomposition.

\paragraph{Context Augmentation and Selection.} 
Context augmentation has been widely explored in video-language generation through memory propagation, retrieval augmentation, and evidence selection~\citep{kim2024you,li2023boosting,yu2023self,xie2025exploring,wu2025event,liu2025task,li2026curvature}. 
Existing methods commonly identify useful context based on semantic similarity, temporal proximity, cross-modal matching, or attention-based relevance. 
However, for information-dense lectures, surface-level relevance does not necessarily reflect prediction utility. 
An earlier definition, assumption, or logical premise may be crucial for the current chapter despite low lexical or visual similarity, while a highly similar segment may simply repeat the current content. 
Our work therefore treats retrieval as candidate construction only, and performs final context selection according to each segment's observable influence on current chapter generation.

\paragraph{Causal Video Reasoning.} 
Causal and counterfactual reasoning has been introduced into video understanding to reduce spurious correlations, mitigate language priors, and model event relations~\citep{xiao2021next,niu2021counterfactual,liu2023cross,chen2025mecd+}. 
Prior studies construct causal video question answering benchmarks, apply counterfactual interventions for bias reduction, or discover event-level causal structures for video reasoning. 
Different from these works, we use lightweight masking and removal interventions to estimate prediction-level influence for long-video chaptering, making our approach closer to intervention-based utility estimation than to causal structure discovery.

\section{Method}
\begin{figure*}[t]
    \centering
    \includegraphics[width=\textwidth]{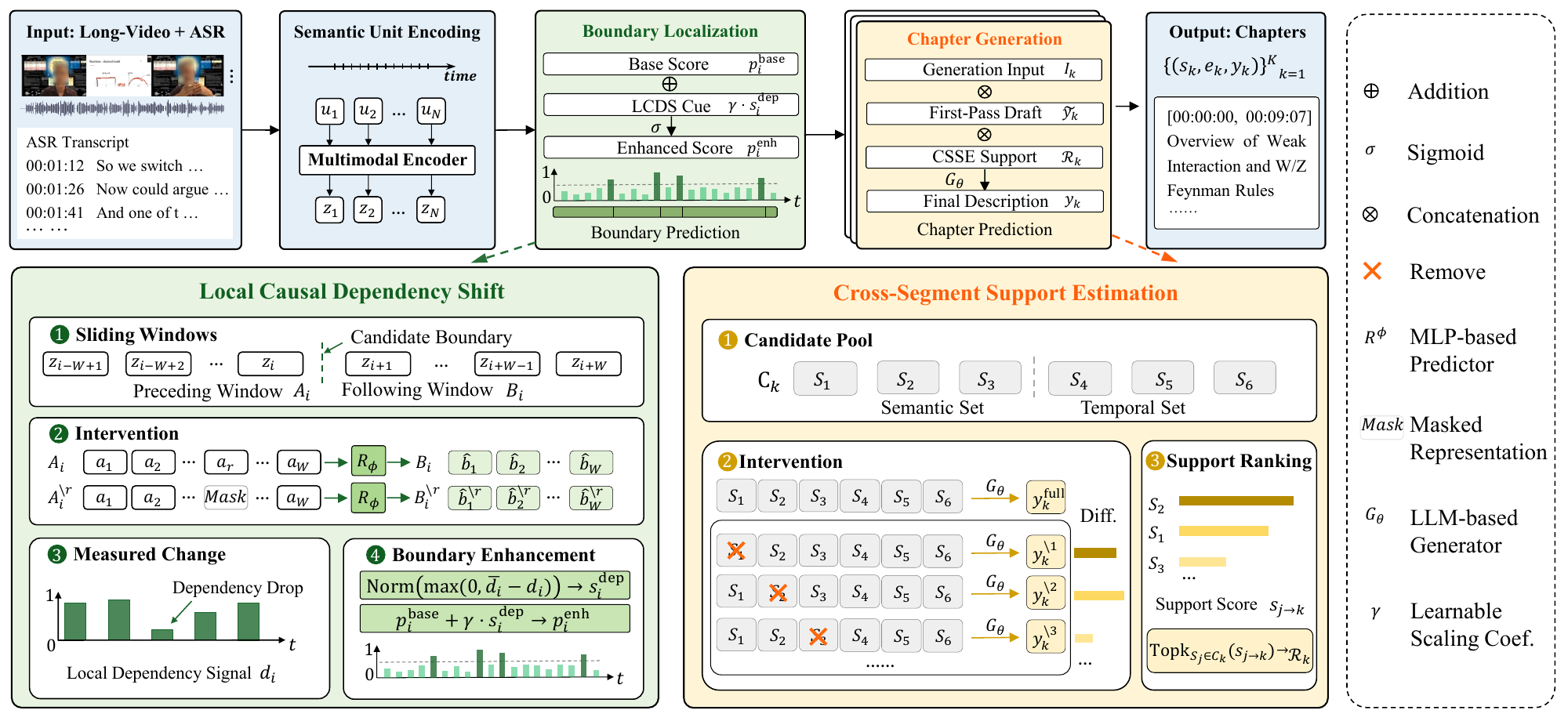}
    \caption{Overview of CausalChapter. Given sentence-level semantic units with aligned visual and ASR information, the framework first predicts chapter boundaries with a segment-then-caption backbone enhanced by Local Dependency Shift (LCDS), which captures dependency drops between adjacent temporal windows. The resulting segments are then described by an LLM-based generator, where Cross-Segment Support Selection (CSSE) reranks historical contexts according to their predictive support for the current chapter. The final output is a sequence of temporally grounded chapter descriptions.}
    \label{fig:framework}
    \vspace{-0.5cm}
\end{figure*}

\subsection{Task Formulation and Overview}

Given a long video $V$, the goal is to generate a temporally grounded chapter sequence $\mathcal{C}=\{(s_k,e_k,y_k)\}_{k=1}^{K}$, where $s_k$ and $e_k$ denote the start and end timestamps of the $k$-th chapter, and $y_k$ denotes its chapter-level description. The predicted chapters are expected to be consecutive, non-overlapping, and semantically coherent.

We propose \textbf{CausalChapter}, a segment-then-caption framework for long-video chapter generation. 
CausalChapter first predicts chapter boundaries and then generates a description for each resulting segment. 
This decomposition is scalable, but it introduces two key challenges, namely boundary error propagation and cross-segment context fragmentation.
Figure~\ref{fig:framework} summarizes the overall pipeline and the two intervention-based modules used to address these challenges.
We address them with two intervention-inspired modules: 
\textbf{Local Dependency Shift} improves boundary localization by measuring drops in predictive dependency between adjacent temporal windows. 
\textbf{Cross-Segment Support Selection} improves chapter generation by selecting historical segments that provide strong predictive support for the current description. 
Here, predictive dependency denotes the observable change in model prediction under masking or removal interventions, and serves as an operational measure of predictive support rather than a claim about real-world causal structure.

\subsection{Segment-then-Caption Backbone}

We instantiate a segment-then-caption backbone over sentence-level semantic units $\mathcal{U}=\{u_i=(x_i,t_i^s,t_i^e)\}_{i=1}^{N}$, where $x_i$ denotes the $i$-th ASR sentence and $t_i^s,t_i^e$ its timestamps.
For each unit, we sample video frames at 1 FPS within $[t_i^s,t_i^e]$ and extract CLIP visual features~\citep{radford2021learning}, and fuse them with the ASR representation to obtain a multimodal semantic-unit representation $z_i$.

Given $z_i$, a boundary classifier $g_{\mathrm{bd}}$ forecasts the probability of a chapter boundary occurring after the $i$-th unit:
\begin{equation}
\label{eq:2}
o_i^{\mathrm{base}}=g_{\mathrm{bd}}(z_i),
p_i^{\mathrm{base}}=\sigma(o_i^{\mathrm{base}})_{c_b},
\end{equation}
where $o_i^{\mathrm{base}}\in\mathbb{R}^{2}$ is boundary logits, $c_b$ is the boundary class index, and $p_i^{\mathrm{base}}$ is the predicted boundary probability. The predicted boundaries are used to divide the video into chapter segments $\mathcal{S}=\{S_k\}_{k=1}^{K}$.

For each segment $S_k$, we construct a generation input $I_k$ from the segment ASR text, a compressed visual representation, an explicit temporal prompt, and a temporal feature pooled from semantic-unit representations within $S_k$. 
The LLM-based generator $G_\theta$ then produces a first-pass chapter description $\tilde{y}_k=G_\theta(I_k)$, where $\theta$ includes trainable parameters such as LoRA adapters and lightweight projection modules. 
This backbone improves scalability, but its boundary prediction mainly relies on local multimodal evidence, and its generation is primarily conditioned on the current segment.

%
%
%

\subsection{Local Dependency Shift}

Smooth topic transitions are difficult to localize from local visual or lexical changes alone, because adjacent units across chapter boundaries may remain semantically coherent. We treat a boundary as positions where the predictive dependency drops between neighboring temporal windows.

For a candidate boundary after $u_i$, we construct a preceding window $A_i=[z_{i-W+1},\ldots,z_i]$ and a following window $B_i=[z_{i+1},\ldots,z_{i+W}]$, where $W$ is the window size. 
A MLP-based dependency predictor $R_\phi$ reconstructs the following window from the preceding one, $\hat{B}_i=R_\phi(A_i)$.
To estimate the contribution of each preceding unit, we mask the $r$-th unit in $A_i$ to obtain $A_i^{\setminus r}$ and predict $\hat{B}_i^{\setminus r}=R_\phi(A_i^{\setminus r})$.

The intervention effect is measured by the representation change in the predicted following window:
\begin{equation}
\label{eq:5}
g_{i,r,j}=D_{\mathrm{rep}}(\hat{b}_{i,j},\hat{b}_{i,j}^{\setminus r}), 
d_i=\frac{1}{W^2}\sum_{r=1}^{W}\sum_{j=1}^{W}g_{i,r,j},
\end{equation}

where $D_{\mathrm{rep}}$ denotes cosine distance, $\hat{b}_{i,j}$ and $\hat{b}_{i,j}^{\setminus r}$ are the $j$-th predicted representations before and after intervention. 
A larger $d_i$ indicates stronger predictive support from the preceding window to the following one, while a smaller $d_i$ indicates weaker temporal dependency.

We then compare $d_i$ with its neighborhood. Let $\mathcal{N}(i)$ denote neighboring candidate positions around $i$ within a fixed local radius, and let $\bar{d}_i=\frac{1}{|\mathcal{N}(i)|}\sum_{q\in\mathcal{N}(i)}d_q$ be the local reference dependency. Since we only care about dependency drops, the final dependency-shift score is
\begin{equation}
\label{eq:6}
s_i^{\mathrm{dep}}=\mathrm{Norm}\big(\max(0,\bar{d}_i-d_i)\big),
\end{equation}
where $\mathrm{Norm}(\cdot)$ denotes min-max normalization over all candidate boundary positions in the video. A larger $s_i^{\mathrm{dep}}$ suggests a stronger local dependency drop and is therefore more likely to indicate a smooth chapter boundary.

Finally, we inject this score into the boundary-class logit, $(o_i^{\mathrm{enh}})_{c_b}
=(o_i^{\mathrm{base}})_{c_b}+\gamma s_i^{\mathrm{dep}},$
where $\gamma$ is a learnable scaling coefficient. 
The enhanced boundary probability is $p_i^{\mathrm{enh}}=\sigma(o_i^{\mathrm{enh}})_{c_b}$. LCDS thus complements the base boundary classifier with an intervention-defined structural cue.

\subsection{Cross-Segment Support Selection}

Independent chapter generation often misses long-range prerequisites such as earlier definitions, assumptions, and problem setups. We therefore estimate the predictive support of historical segments through removal interventions.
For each segment $S_k$, we build a segment representation $h_k=E_{\mathrm{seg}}(S_k,\tilde{y}_k)$ using the segment content, temporal information, segmentation-stage features, and first-pass description. 
To control computation, we first construct a compact candidate context set:
\begin{equation}
\label{eq:8}
\begin{aligned}
C_k &= C_k^{\mathrm{near}}\cup C_k^{\mathrm{sem}},\\
C_k^{\mathrm{sem}} &= \mathrm{TopM}_{j<k}\mathrm{Sim}(h_k,h_j).
\end{aligned}
\end{equation}
where $C_k^{\mathrm{near}}$ contains temporally neighboring historical segments, $C_k^{\mathrm{sem}}$ contains semantically retrieved segments, and $M$ is a small retrieval budget. This stage only narrows the search space and does not determine the final contexts.

Given $C_k$, the generator first produces a reference sequence with all candidate contexts,
$y_k^{\mathrm{ref}}=G_\theta(I_k,\tilde{y}_k,C_k)$.
For each candidate segment $S_j\in C_k$, let $C_k^{-j}=C_k\setminus\{S_j\}$. We compare the teacher-forced output distributions under $C_k$ and $C_k^{-j}$ over the same reference prefix. Specifically, define
\begin{equation}
p_t(C)=p_\theta(\cdot \mid y_{k,<t}^{\mathrm{ref}}, I_k,\tilde{y}_k,C).
\end{equation}
The output-difference function is then
\begin{equation}
\label{eq:dout}
D_{\mathrm{out}}(C_k,C_k^{-j})
=\frac{1}{T}\sum_{t=1}^{T}
D_{\mathrm{KL}}\!\big(p_t(C_k)\|p_t(C_k^{-j})\big).
\end{equation}
Here, $T$ is the number of evaluated output positions. A larger divergence means that removing $S_j$ changes the model's predictive distribution more strongly, so $S_j$ is considered to provide stronger predictive support for the current chapter. We define its support score as
\begin{equation}
\label{eq:11}
s_{j\rightarrow k}=D_{\mathrm{out}}(C_k,C_k^{-j}),
\end{equation}
and analyze alternative implementations of $D_{\mathrm{out}}$ in Appendix~\ref{app:csse_sensitivity}.

The top-ranked contexts are then selected as $\mathcal{R}_k=\mathrm{TopK}_{S_j\in C_k}(s_{j\rightarrow k})$, and the final chapter description is generated as $y_k=G_\theta(I_k,\tilde{y}_k,\mathcal{R}_k)$. 
This shifts context selection from similarity-driven matching to intervention-driven support estimation. Since interventions are performed only on the compact candidate set $C_k$, the additional cost scales with the candidate size rather than the total number of video segments.

\subsection{Training Objective and Inference}

The boundary module is trained with supervised boundary labels using
\begin{equation}
\label{eq:13}
\mathcal{L}_{\mathrm{bd}}
=-\sum_i \log p_i^{\mathrm{enh}}(b_i^\ast),
\end{equation}
where $b_i^\ast$ is the ground-truth boundary label after $u_i$. The dependency predictor is trained to reconstruct the following window with $\mathcal{L}_{\mathrm{rec}}=\sum_i D_{\mathrm{rec}}(\hat{B}_i,B_i)$.

The generator is trained with the standard auto-regressive objective:
\begin{equation}
\label{eq:14}
\mathcal{L}_{\mathrm{gen}}
=-\sum_{k}\sum_{t}
\log p_\theta(y_{k,t}^{\ast}\mid y_{k,<t}^{\ast}, I_k,\mathcal{R}_k),
\end{equation}
where $y_k^{\ast}$ is the ground-truth chapter description, together with a timestamp reconstruction loss
\begin{equation}
\label{eq:15}
\mathcal{L}_{\mathrm{time}}
=-\sum_{t\in\Omega_{\mathrm{time}}}
\log p_\theta(y_t \mid y_{<t}, I_k).
\end{equation}

The final objective is
\begin{equation}
\label{eq:16}
\mathcal{L}
=\mathcal{L}_{\mathrm{bd}}+
\mathcal{L}_{\mathrm{rec}}+
\mathcal{L}_{\mathrm{gen}}+
\mathcal{L}_{\mathrm{time}},
\end{equation}

At inference time, CausalChapter first encodes semantic units and predicts enhanced chapter boundaries, which define chapter segments. 
The generator then produces first-pass descriptions, retrieves candidate historical contexts, estimates their support scores through removal interventions, and generates final descriptions with the top supportive contexts. 
The final output is the chapter sequence $\mathcal{C}=\{(s_k,e_k,y_k)\}_{k=1}^{K}$.

\section{Experiments}

\subsection{Experimental Setup}
\paragraph{Datasets.}
We evaluate CausalChapter on two long-video chaptering benchmarks. 
The primary benchmark is \textbf{AVLecture}~\citep{gupta2023unsupervised}, which contains long lecture videos with ASR transcripts, OCR outputs, visual content, and human-annotated topic boundaries. 
Since AVLecture does not provide chapter-level descriptions, we augment each ground-truth segment with a human-verified LLM-assisted chapter description. 
All models are trained and evaluated on the same augmented references. 
We use 245 videos for training, 35 for validation, and 70 for testing. 
We further evaluate generalization on \textbf{VidChapters-7M}~\citep{yang2023vidchapters}, which provides user-annotated chapter boundaries and titles. 
Details of the annotation protocol and annotation-sensitivity analysis are provided in Appendix~\ref{app:annotation}.




\paragraph{Metrics.}
We evaluate both chapter localization and description generation.
For localization, we report F1@30 and tIoU on AVLecture, and follow prior work to report F1 and tIoU on VidChapters-7M.
We additionally use exact-boundary F1 in ablation analyses to expose fine-grained differences among boundary variants.
For generation, we report CIDEr~\citep{vedantam2015cider} on both datasets.
On AVLecture, we report SODA\_c~\citep{fujita2020soda}, computed with SODA type-c using IoU-weighted METEOR matching.
On VidChapters-7M, we follow the Chapter-Llama evaluation and report its benchmark notation SODA.
\paragraph{Baselines.}
We compare CausalChapter with representative dense video captioning methods, including PDVC~\citep{wang2021end} and Vid2Seq~\citep{yang2023vid2seq}, 
as well as long-video LLM and video chaptering baselines, including VTimeLLM~\citep{huang2024vtimellm} and Chapter-Llama~\citep{ventura2025chapter}. 
%
We also include closed-source LLMs, GPT-4o~\citep{hurst2024gpt}, Gemini-2.5-Pro and Gemini-2.0-Flash~\citep{team2023gemini}, as reference systems.  
All trainable baselines are adapted to the same input setting and evaluated with the same references whenever applicable.

\paragraph{Implementation.}
For state-of-the-art comparison, we evaluate both Qwen2.5-7B~\citep{qwen2025qwen25technicalreport} and LLaMA-3.1-8B~\citep{grattafiori2024llama3herdmodels} on AVLecture and VidChapters-7M.
For controlled ablations, we use Qwen2.5-3B as the generation backbone and keep it fixed across all variants. 
%
All trainable LLMs are adapted with LoRA, where we set the rank and scaling factor to $(r,\alpha)=(32,64)$ for Qwen2.5-7B and $(r,\alpha)=(8,16)$ for Qwen2.5-3B.
Models are trained with AdamW using a learning rate of $5 \times 10^{-5}$, batch size $1$, and maximum input length $8192$.
For prompting, all methods use the same segment-level instruction template, which includes the segment ASR, visual summary, timestamp prompt, and optional retrieved contexts. 
For CSSE, we construct candidate contexts from temporal neighbors and semantic retrieval, and select the top-$K=5$ segments. For LCDS, we set the window size to $W=5$ for interventional dependency modeling. 
We use deterministic decoding with temperature $0$ for intervention scoring to ensure that output changes are attributable to context removal, and use the same decoding setting across all compared variants. 
%
Experiments are conducted on a Debian GNU/Linux 12 server equipped with a single NVIDIA H800 PCIe GPU with $80$ GB memory.
Additional efficiency and scalability analyses are reported in Appendix~\ref{app:efficiency}.
\begin{table}[t]
\centering
\scriptsize
\setlength{\tabcolsep}{4pt}
\renewcommand{\arraystretch}{1.08}

\vspace{-0.4em}
\resizebox{\linewidth}{!}{
\begin{tabular}{@{}l l c c c c@{}}
\toprule
\textbf{Type} & \textbf{Method}
& \textbf{CIDEr} & \textbf{SODA\_c} & \textbf{F1@30} & \textbf{tIoU} \\
\midrule
\multirow{2}{*}{DVC}
 & PDVC & 6.11 & -- & 9.07 & 56.44 \\
 & Vid2Seq\textsuperscript{T5} & 61.22 & 9.79 & 9.84 & 53.77 \\
\midrule
\multirow{5}{*}{Chap.}
 & VTimeLLM\textsuperscript{Vicuna, *} & 0.02 & 3.04 & 47.53 & 46.72 \\
 & Chapter-Llama\textsuperscript{LLaMA} & 99.78 & 8.15 & 63.93 & 62.18 \\
 & Chapter-Llama\textsuperscript{Qwen} & 95.36 & 6.91 & 57.73 & 59.62 \\
 & CausalChapter\textsuperscript{LLaMA} & 110.12 & 11.73 & \textbf{72.97} & \textbf{69.95} \\
 & CausalChapter\textsuperscript{Qwen} & \textbf{116.88} & \textbf{13.80} & \textbf{72.97} & \textbf{69.95} \\
\midrule
\multirow{3}{*}{LLM}
 & GPT-4o & 1.82 & 0.08 & 24.89 & 35.23 \\
 & Gemini-2.5-Pro & 2.40 & 1.64 & 26.72 & 36.12 \\
 & Gemini-2.0-Flash & 0.28 & 0.18 & 11.71 & 13.54 \\
\bottomrule
\end{tabular}
}
\vspace{-0.4em}
\caption*{{(a) AVLecture}}
\vspace{0.4em}

\resizebox{\linewidth}{!}{
\begin{tabular}{@{}l l c c c c@{}}
\toprule
\textbf{Type} & \textbf{Method}
& \textbf{CIDEr} & \textbf{SODA} & \textbf{F1} & \textbf{tIoU} \\
\midrule
DVC
 & Vid2Seq\textsuperscript{T5} & 55.8 & 11.6 & 26.7 & 58.6 \\
\midrule
\multirow{3}{*}{Chap.}
 & Chapter-Llama\textsuperscript{LLaMA} & 100.9 & \textbf{19.3} & 45.3 & 71.8 \\
 & CausalChapter\textsuperscript{LLaMA} & 101.5 & 18.5 & \textbf{46.2} & \textbf{72.0} \\
 & CausalChapter\textsuperscript{Qwen} & \textbf{109.2} & 18.8 & \textbf{46.2} & \textbf{72.0} \\
\midrule
\multirow{2}{*}{LLM}
 & GPT-4o & 51.0 & 8.1 & 37.6 & 68.0 \\
 & Gemini-2.0-Flash & 69.7 & 11.4 & 40.2 & 69.3 \\
\bottomrule
\end{tabular}
}
\vspace{-0.4em}
\caption*{{(b) VidChapters-7M}}
\caption{
Comparison of video chaptering methods on AVLecture and VidChapters-7M.
For CausalChapter, localization scores come from the shared boundary-localization stage, while generation scores use the indicated backbone.
Superscripts denote backbones:
\textsuperscript{T5} T5,
\textsuperscript{Vicuna} Vicuna-7B,
\textsuperscript{LLaMA} LLaMA-3.1-8B, and
\textsuperscript{Qwen} Qwen2.5-7B.
Methods without superscripts are proprietary or do not report a backbone; \textsuperscript{*} marks VTimeLLM not fine-tuned on AVLecture.
}
\label{tab:main_comparison}
\vspace{-1.5em}
\end{table}

\begin{table*}[t]
  \centering
  \small
  \setlength{\tabcolsep}{4pt}
  \begin{tabular}{lcccccc}
    \hline
    \textbf{Method} &  \textbf{CIDEr} & \textbf{METEOR} & \textbf{SODA\_c} & \textbf{F1}  & \textbf{BS@30} & \textbf{tIoU} \\
    \hline
    Backbone  & 88.39 & 16.83 & 10.84 & 52.95 & 59.61 & 66.36 \\
    + LCDS  & 91.36 & 17.37 & 11.90 & \textbf{56.72}  & \textbf{63.02} & \textbf{69.95} \\
    + CSSE  & 98.40 & 17.15 & 12.63 & 52.95  & 59.61 & 66.36 \\
    Full  & \textbf{104.59} & \textbf{17.87} & \textbf{12.73} & \textbf{56.72}  & \textbf{63.02} & \textbf{69.95} \\
    \hline
  \end{tabular}
  \caption{\label{tab:ablation}
    Ablation studies on AVLecture. All variants use predicted boundaries. LCDS denotes Local Causal Dependency Shift, CSSE denotes Cross-Segment Causal Support Estimation. 
  }
  \vspace{-1em}
\end{table*}

\begin{table}[t]
  \centering
  \small
  \setlength{\tabcolsep}{4pt}
  \begin{tabular}{lcccc}
  
    \hline
    \textbf{Method} & \textbf{F1} & \textbf{F1@30}   & \textbf{BS@30} & \textbf{tIoU} \\
    \hline
    Baseline & 52.95 &71.28&	59.61&	66.36  \\
    Sim. Drop & 54.51&	72.41&	61.80&	65.68 \\
    CL Loss & 53.28 &	70.91&	59.75&	65.23 \\
    Ours & \textbf{56.72}&	\textbf{72.97}&	\textbf{63.02}&\textbf{	69.95} \\
    \hline
  \end{tabular}
  \caption{\label{tab:seg}
        Analysis of LCDS on AVLecture. All methods use the same Qwen2.5-3B backbone.
  }
\end{table}
\subsection{Main Result}
\label{sec:sota}
We compare CausalChapter with dense video captioning methods, long-video LLM baselines, video chaptering models, and closed-source LLMs under zero-shot prompting. 
Tab.~\ref{tab:main_comparison} reports results on the augmented AVLecture and VidChapters-7M benchmarks.

\paragraph{Results on AVLecture.}
CausalChapter performs best overall among trainable models, substantially outperforming the DVC baselines PDVC~\citep{wang2021end} and Vid2Seq~\citep{yang2023vid2seq}. 
Compared with the strongest fine-tuned chaptering baseline, Chapter-Llama~\citep{ventura2025chapter}, it improves CIDEr from 99.78 to 110.12 with LLaMA-3.1-8B and from 95.36 to 116.88 with Qwen2.5-7B; F1@30 and tIoU also increase from 63.93\% and 62.18\% to 72.97\% and 69.95\%, respectively. 
These gains demonstrate improvements in both description quality and boundary localization. 
Closed-source LLMs under zero-shot prompting obtain lower automatic scores; Appendix~\ref{app:closed_llm_semantic} provides a complementary semantic evaluation, and Appendix~\ref{app:qualitative} provides a qualitative comparison with Chapter-Llama.

\paragraph{Results on VidChapters-7M.}
Tab. \ref{tab:main_comparison}(b) shows that our method also generalizes to VidChapters-7M, which contains more diverse open-domain videos with user-annotated chapter boundaries and titles. 
Despite using a compact backbone, CausalChapter achieves competitive or better performance than long-video LLM baselines on both generation and localization metrics. 
This suggests that the proposed dependency-enhanced segment-then-caption framework is not specific to lecture videos and can transfer to broader long-video chaptering scenarios.

\subsection{Ablation Studies}
\label{sec:ablation}


We ablate CausalChapter on AVLecture using the same Qwen2.5-3B backbone and training data. \textit{Backbone} is the base segment-then-caption model; \textit{+LCDS} and \textit{+CSSE} add the corresponding localization and generation modules, while \textit{Full} further includes temporal-aware training.

As shown in Table~\ref{tab:ablation}, LCDS improves F1, BS@30, and tIoU from 52.95\%, 59.61\%, and 66.36\% to 56.72\%, 63.02\%, and 69.95\%, respectively. CSSE leaves boundary predictions unchanged while increasing CIDEr from 88.39 to 98.40 and SODA\_c from 10.84 to 12.63. Their complementary effects yield the full model's best CIDEr, METEOR, and SODA\_c scores of 104.59, 17.87, and 12.73.

\subsection{Analysis of Interventional Dependency Modeling}
\label{sec:analysis}


We further analyze whether intervention-defined dependency provides more useful signals than simpler relevance-based alternatives. 
%
Table~\ref{tab:seg} compares LCDS with representation-similarity drops (\textit{Sim. Drop}) and a contrastive objective (\textit{CL Loss}). Similarity drops improve several threshold-based metrics but reduce tIoU, while contrastive learning gives only marginal gains. Our dependency score performs best on all metrics, increasing F1 from 52.95\% to 56.72\% and tIoU from 66.36\% to 69.95\%, supporting predictive dependency as a stronger cue for smooth boundaries.


\begin{table}[t]
  \centering
  \small
  \setlength{\tabcolsep}{7pt}
  \begin{tabular}{llcccc}
    \hline
    \textbf{Method}  & \textbf{CIDEr} & \textbf{METEOR} & \textbf{SODA\_c} \\
    \hline
    Baseline & 88.39 & 16.83 & 10.84 \\
    Previous-$K$ & 90.85 & 16.70 & 11.32 \\
    Sim. Top-$K$  & 95.97 & 16.79 & 11.70 \\
    LLM Scoring  & 96.11 & 17.44 & 11.53 \\
    Ours & \textbf{104.59} & \textbf{17.87} & \textbf{12.73} \\
    \hline
  \end{tabular}
  \caption{\label{tab:context_selection}
    Analysis of cross-segment context selection on AVLecture. All methods use Qwen2.5-3B as backbone.
  }
  \vspace{-0.5cm}
\end{table}

Table~\ref{tab:context_selection} compares CSSE with temporal proximity (\textit{Previous-$K$}), semantic retrieval (\textit{Sim. Top-$K$}), and LLM-estimated relevance using the same backbone and boundaries. CSSE performs best on all generation metrics, improving CIDEr from 88.39 to 104.59 and SODA\_c from 10.84 to 12.73, showing the advantage of measuring a context's influence on current generation.

\begin{figure}[t]
    \centering
    \includegraphics[width=\linewidth]{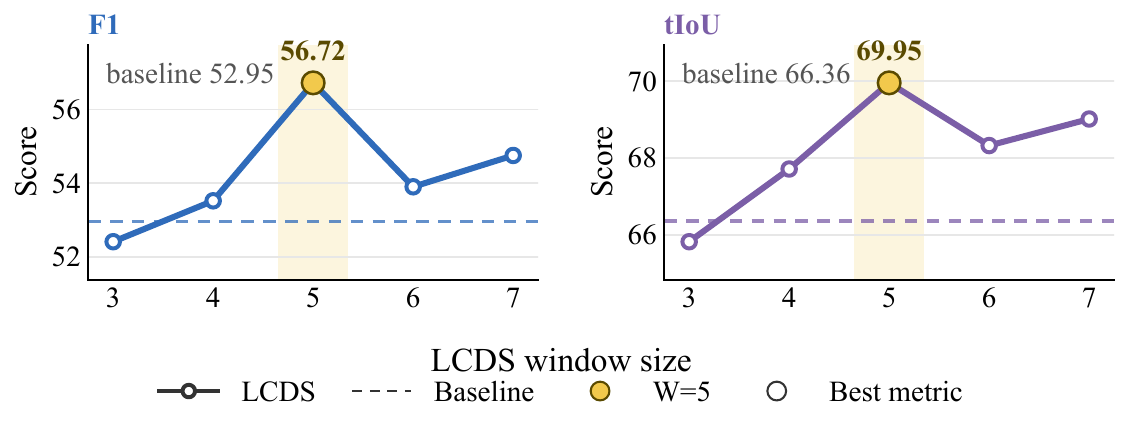}
    \caption{
Window-size sensitivity of LCDS on AVLecture.
}
    \label{fig:window_size}
    \vspace{-0.5cm}
\end{figure}

\paragraph{Window-size sensitivity.}
We further analyze the effect of the LCDS window size \(W\).
As shown in Figure~\ref{fig:window_size}, \(W=5\) provides the most balanced performance across boundary-oriented metrics, improving F1, BS@30, and tIoU over the backbone while maintaining competitive F1@30.
This suggests that a moderate local window captures sufficient cross-boundary dependency changes without introducing excessive neighboring noise.

\begin{figure}
    \centering
    \includegraphics[width=\linewidth]{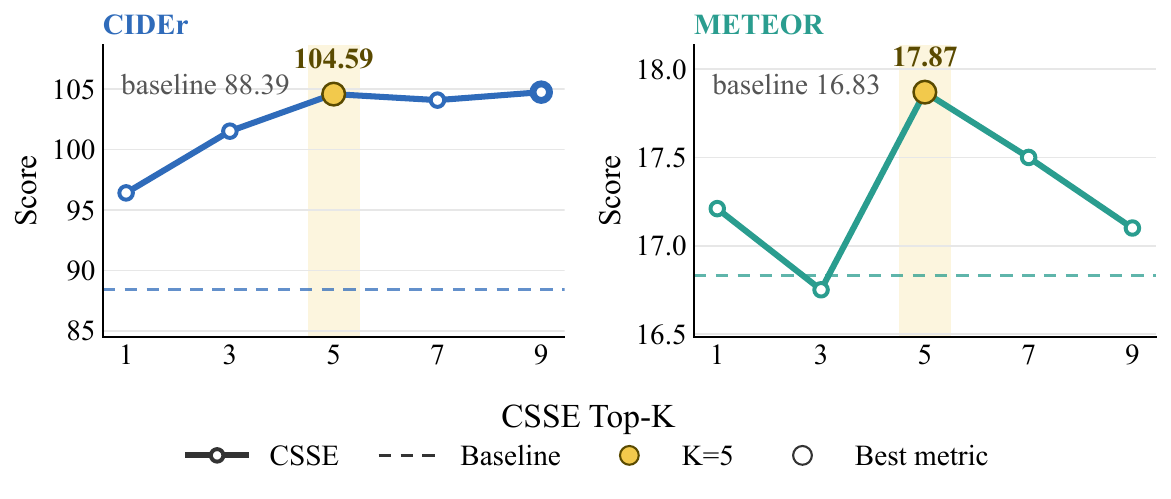}
    \caption{
Top-$K$ sensitivity of CSSE on AVLecture.
} 
    \label{fig:topk_hyperparam}
\vspace{-0.5cm}
\end{figure}

\paragraph{Top-$K$ sensitivity.}
We further analyze the effect of the number of supportive contexts selected by CSSE.
As shown in Figure~\ref{fig:topk_hyperparam}, increasing Top-$K$ generally improves generation quality over the backbone, indicating that cross-segment supportive contexts provide useful complementary information for chapter description generation.
The setting \(K=5\) achieves the best METEOR score and a strong CIDEr score, improving CIDEr from 88.39 to 104.59 and METEOR from 16.83 to 17.87.
Therefore, we use \(K=5\) as a balanced setting in our main experiments.
\section{Conclusion}

We presented \textbf{CausalChapter}, an interventional dependency modeling framework for long-video chaptering. 
To address the boundary error propagation and cross-segment context fragmentation introduced by segment-level generation, CausalChapter estimates intervention-defined predictive dependencies as task-oriented support signals. 
Specifically, LCDS captures local dependency drops between adjacent temporal windows to provide auxiliary evidence for smooth chapter boundaries, while CSSE selects cross-segment contexts according to their intervention-defined support for current chapter generation. 
Experiments on AVLecture and VidChapters-7M show that CausalChapter improves temporal localization and description quality on AVLecture, while remaining competitive on VidChapters-7M.
These results highlight intervention-defined dependency as a useful signal for scalable and coherent long-video chaptering.

\section*{Limitations}

CausalChapter has several limitations. 
First, intervention-defined dependency should be interpreted as prediction-level influence rather than real-world causal discovery. 
Our goal is not to recover causal relations among video events or chapters, but to identify semantic units or context segments that affect boundary prediction and chapter-level generation under controlled masking or removal interventions.  
Second, our augmented AVLecture benchmark uses human-verified LLM-assisted chapter descriptions, which may inherit some stylistic regularities from the annotation pipeline. 
To reduce this effect, all compared methods are trained and evaluated with the same augmented references, and our annotation-sensitivity analysis examines alternative annotation models and styles. 
Third, CSSE introduces additional inference cost because it estimates context support through removal interventions. 
We control this cost by applying interventions only to a compact candidate set constructed from temporal neighbors and semantic retrieval, rather than to all segments or tokens. 
Further acceleration of support estimation is an interesting direction for future work.

\section*{Acknowledgments}

This work was supported by the National Natural Science
Foundation of China (62437001, 62506040 and 62402051) and the Fundamental Research Funds for the Central Universities  (2253500001). 

\bibliography{custom}

\clearpage
\appendix
\section{Method Details}

\subsection{Segmentation Head Architecture}
\label{app:segmentation_head}

The main text abstracts the boundary predictor as $g_{\mathrm{bd}}(z_i)$. In practice, our segmentation head follows the general design of recent multimodal video topic segmentation models: it uses sentence-aligned clips as basic units, projects visual and textual clip features into a shared space, performs middle-fusion across modalities, and then predicts whether each unit is a topic boundary with a lightweight binary classifier.

Concretely, for the $i$-th semantic unit, we first obtain a visual clip representation and a text representation from sampled video frames and the corresponding ASR sentence, where $E_v$ and $E_t$ denote the visual and textual encoders, respectively. After projection into the same hidden dimension by the trainable projection matrices $W_v$ and $W_t$, the two modalities are fused by a small stack of multimodal fusion layers, denoted by $\mathrm{MFL}$, to produce updated visual and textual states. We then concatenate the fused modality-specific states into the multimodal unit representation $z_i$, which is used by the segmentation head:
\begin{equation}
v_i = W_v E_v(c_i^v), \qquad t_i = W_t E_t(c_i^t),
\end{equation}
\begin{equation}
h_i^v, h_i^t = \mathrm{MFL}(v_i, t_i), \qquad z_i = [h_i^v; h_i^t],
\end{equation}
\begin{equation}
o_i^{\mathrm{base}} = g_{\mathrm{bd}}(z_i) = W_p z_i + b_p.
\end{equation}

Here, $W_p$ and $b_p$ denote the predictor weight matrix and bias term of the final binary classifier. This design keeps the segmentation head lightweight while still allowing cross-modal interaction before boundary prediction. Relative to a late-fusion design, the middle-fusion structure better exposes cross-modal cues such as transcript transitions, slide changes, and visual context shifts to the boundary classifier. In our implementation, the segmentation head is therefore best viewed as a multimodal fusion block followed by a linear classifier that outputs the boundary logits in Eq.~\ref{eq:2}.

\section{Dataset Construction and Annotation}

\subsection{AVLecture Benchmark}
\label{app:annotation}

AVLecture is a long-form instructional-video benchmark introduced to audio-visual lecture segmentation. The full AVLecture collection contains 86 courses with over 2,350 lectures and a total duration of roughly 2,200 hours, covering a broad range of STEM subjects. Each course provides video lectures together with aligned ASR transcripts and OCR signals, and many courses also include auxiliary educational resources such as lecture notes, slides, and assignments. Among the 86 courses, a 15-course subset with 350 lectures is annotated with temporal segmentation boundaries and serves as the standard benchmark for lecture segmentation. In our work, we adopt this segmented subset as the boundary-localization foundation, and use the 245/35/70 train/validation/test split described in the main text. We further extend these lectures with chapter-level reference descriptions so that the same benchmark can support evaluation of both chapter localization and chapter description generation.


\paragraph{LLM-Assisted Annotation Pipeline.}

For each video, we use GPT-4o~\citep{hurst2024gpt} as the annotation model. The input includes the full video transcript, frame-level captions extracted every 10 seconds, and the start and end timestamps of all ground-truth segments. The model is instructed to output a JSON object, where each segment is paired with a concise chapter-level description. Each description is expected to summarize the main topic of the corresponding segment, remain faithful to the transcript and frame captions, and distinguish the segment from adjacent chapters. When the transcript and frame captions exceed the input budget, we truncate the input while preserving the segment timestamps and the local transcript/caption context around each target segment.

\paragraph{Human Verification and Revision.}
After GPT-4o-assisted annotation, all segment descriptions are manually checked and revised. We correct descriptions that are overly generic, unsupported by the transcript or frame captions, inconsistent with the segment boundary, or redundant with adjacent chapters. For segments whose content is distributed across multiple stages of the lecture, we manually improve or combine the generated descriptions to better reflect the complete chapter-level semantics. We also normalize the JSON format, description length, and writing style across videos. The final references are fixed before training and evaluation, and all compared methods use the same augmented references, ensuring that the evaluation remains internally consistent.

The verification was conducted by 10 annotators. Each video and all of its chapter descriptions were reviewed by a group of three annotators. When one annotator revised a description, the revised label was circulated to the other two annotators and further refined until all three accepted the final version. We therefore record revision history and final consensus rather than a chance-corrected agreement coefficient. Among the 350 videos, 308 videos (88.0\%) required no revision, 39 videos (11.1\%) required one revision round, and 3 videos (0.9\%) required two revision rounds. All final labels were accepted by the assigned annotator group.

\begin{table*}[t]
  \centering
  \small
  \setlength{\tabcolsep}{3pt}
  \renewcommand{\arraystretch}{1.08}
  \begin{tabular}{
    >{\raggedright\arraybackslash\hyphenpenalty=10000\exhyphenpenalty=10000}p{0.17\textwidth}
    >{\raggedright\arraybackslash\hyphenpenalty=10000\exhyphenpenalty=10000}p{0.26\textwidth}
    >{\raggedright\arraybackslash\hyphenpenalty=10000\exhyphenpenalty=10000}p{0.31\textwidth}
    >{\raggedright\arraybackslash\hyphenpenalty=10000\exhyphenpenalty=10000}p{0.205\textwidth}}
    \toprule
    \textbf{Issue} & \textbf{LLM draft} & \textbf{Human-revised label} & \textbf{Reason} \\
    \midrule
    Cross-boundary content
    & Force range dependence on mediator mass via decay laws
    & Force range dependence on mediator mass
    & ``Decay laws'' belongs to the following chapter. \\
    Overly general description
    & Conservation laws in elastic collisions
    & Linearizing one-dimensional elastic collisions via relative velocity
    & The draft omitted the specific derivation in the current chapter. \\
    \bottomrule
  \end{tabular}
  \caption{Representative human revisions in the augmented AVLecture benchmark.}
  \label{tab:annotation_revision_examples}
\end{table*}

Table~\ref{tab:annotation_revision_examples} gives representative revision examples, illustrating how human annotators remove cross-boundary content and make chapter descriptions more specific to the current segment.

\paragraph{Annotation Sensitivity.}
To examine whether model comparisons depend on a particular annotation model or writing style, we construct four AVLecture reference sets by varying the annotation model and description format: GPT-4o phrase-style references, GPT-4o sentence-style references, Claude phrase-style references, and Claude sentence-style references. For each reference set, models are retrained and evaluated with the same split, model configuration, training procedure, and evaluation code; only the training and evaluation references are changed. Table~\ref{tab:annotation_sensitivity} shows that CausalChapter improves over Chapter-Llama across all four reference settings, suggesting that the gains are not tied to a single annotation style.

\begin{table*}[t]
  \centering
  \small
  \setlength{\tabcolsep}{3pt}
  \resizebox{\textwidth}{!}{%
  \begin{tabular}{lcccccccccc}
    \toprule
    \textbf{Method} &
    \textbf{R1 C} & \textbf{R1 S} &
    \textbf{R2 C} & \textbf{R2 S} &
    \textbf{R3 C} & \textbf{R3 S} &
    \textbf{R4 C} & \textbf{R4 S} &
    \textbf{Avg. C} & \textbf{Avg. S} \\
    \midrule
    Chapter-Llama\textsuperscript{LLaMA} & 99.78 & 8.15 & 101.26 & 8.41 & 107.25 & 10.32 & 104.74 & 10.13 & 103.26 & 9.25 \\
    CausalChapter\textsuperscript{LLaMA} & 110.12 & 11.73 & 116.12 & 14.13 & 115.63 & 12.67 & 128.19 & 14.34 & 117.52 & 13.22 \\
    CausalChapter\textsuperscript{Qwen} & 116.88 & 13.80 & 138.99 & 15.21 & 121.74 & 14.63 & 145.24 & 15.83 & 130.71 & 14.87 \\
    \bottomrule
  \end{tabular}%
  }
  \caption{Annotation-sensitivity results on AVLecture. R1/R2 use GPT-4o references in phrase/sentence styles, and R3/R4 use Claude references in phrase/sentence styles. C and S denote CIDEr and SODA\_c, respectively.}
  \label{tab:annotation_sensitivity}
\end{table*}

\section{Additional Experimental Results}

\subsection{Evaluation Metrics}
\label{app:more_metrics}

For completeness, we briefly summarize the metrics used in the main paper and the appendix tables. \textbf{CIDEr} evaluates how well a generated chapter description matches the reference description using consensus-based $n$-gram similarity, with higher scores indicating better agreement with human-written references. \textbf{METEOR} measures generation quality through unigram alignment with stemming and synonym matching. \textbf{SODA\_c} jointly evaluates temporal localization and description quality through temporally ordered matching; on AVLecture, we use SODA type-c with IoU-weighted METEOR matching, as specified in the main text.

For boundary localization, \textbf{F1} is the harmonic mean of precision and recall, and evaluates the overall quality of predicted topic boundaries according to how well they match ground-truth boundaries. \textbf{BS@30} (Boundaries at 30 seconds) measures whether a predicted boundary falls within a 30-second tolerance window around a ground-truth boundary, thus reflecting boundary-detection accuracy under a fixed temporal tolerance. \textbf{tIoU} reports the temporal intersection-over-union between predicted and ground-truth segments, measuring how accurately the predicted chapter partition overlaps with the reference segmentation. \textbf{F1@30} applies the F1 criterion under the same 30-second matching window, and therefore captures both boundary accuracy and temporal tolerance.

\subsection{Backbone Scaling and Complete Baseline Results}
\label{app:backbone_scaling}



\paragraph{Overview.}
The main experiments use different backbone sizes for different purposes: we use larger backbones for state-of-the-art comparison, and a smaller Qwen2.5-3B backbone for ablation studies to control experimental cost while keeping variants comparable. This appendix reports the complete results behind these choices. We first present backbone scaling results for CausalChapter across the LLaMA and Qwen2.5 families, and then provide the full LLM-based baseline and closed-source LLM reference results on AVLecture. These supplementary results verify that the gains of CausalChapter are not tied to a single model family or scale.
\begin{table}[t]
  \centering
  \small
  \setlength{\tabcolsep}{5pt}
  \resizebox{\linewidth}{!}{%
  \begin{tabular}{lcccc}
    		\toprule
    		\textbf{Backbone} & \textbf{temporal-aware} & \textbf{CIDEr} & \textbf{METEOR} 
            \\
    \midrule
    LLaMA-3.2-1B & w/o& 78.58 & 10.58 
    \\
    LLaMA-3.2-1B  & with & 83.16 & 13.05 
    \\
    LLaMA-3.2-3B & w/o & 93.40 & 17.15 
    \\
    LLaMA-3.2-3B & with & 98.45 & 17.45 
    \\
    LLaMA-3.1-8B & w/o & 98.38 & 19.08 
    \\
    LLaMA-3.1-8B & with & 110.12 & 19.41 
    \\
    \midrule
    Qwen2.5-0.5B & w/o & 84.56 & 10.38 
    \\
    Qwen2.5-0.5B & with & 86.48 & 13.41 
    \\
    Qwen2.5-1.5B & w/o & 90.57 & 16.19 
    \\
    Qwen2.5-1.5B & with& 97.94 & 16.33 
    \\
    Qwen2.5-3B & w/o & 97.64 & 17.73 
    \\
    Qwen2.5-3B & with& 104.59 & 17.87 
    \\
    Qwen2.5-7B & w/o & 113.98 & 18.24 
    \\
    Qwen2.5-7B & with & 116.88 & 20.43 
    \\
    \bottomrule
  \end{tabular}%
  }
  \caption{\label{tab:app_backbone_scaling}
    Backbone scaling results of CausalChapter on AVLecture. Temporal-aware optimization further injects timestamp information and segmentation-stage temporal features into the generator.
  }
\end{table}

Table~\ref{tab:app_backbone_scaling} shows two consistent trends. First, stronger backbones generally lead to better chapter-level generation quality, especially on CIDEr. Second, temporal-aware optimization improves most corresponding settings across both LLaMA and Qwen2.5 families. These results suggest that the proposed framework benefits from model scaling, while the temporal-aware optimization provides additional gains beyond simply increasing backbone size.
\begin{table*}[t]
  \centering
  \small
  \setlength{\tabcolsep}{4pt}
  \resizebox{\textwidth}{!}{%
  \begin{tabular}{lllrrrrrrr}
    		\toprule
    		\textbf{Method} & \textbf{Backbone} & \textbf{Type} & \textbf{CIDEr} & \textbf{METEOR} 
            & \textbf{F1} & \textbf{F1@30} & \textbf{BS@30} & \textbf{tIoU} \\
    \midrule
    VTimeLLM & ChatGLM3-6B & ZS & 0.12 & 1.79 
    & 32.46 & 47.66 & 45.78 & 48.21 \\
    VTimeLLM & Vicuna-7B & ZS & 0.02 & 2.04 
    & 32.25 & 47.53 & 50.97 & 46.72 \\
    VTimeLLM & Vicuna-7B & FT & -- & 0.59 
    & 9.22 & 20.71 & 37.38 & 27.88 \\
    \midrule
    Chapter-Llama & LLaMA-3.2-1B & ZS & 14.77 & 8.36 
    & 3.51 & 8.19 & 8.19 & 19.14 \\
    Chapter-Llama & LLaMA-3.2-1B & FT & 68.53 & 15.67 
    & 48.18 & 63.36 & 63.36 & 59.30 \\
    Chapter-Llama & LLaMA-3.2-3B & ZS & 38.77 & 12.79 
    & 4.39 & 9.12 & 9.12 & 23.58 \\
    Chapter-Llama & LLaMA-3.2-3B & FT & 86.07 & 19.40 
    & 47.53 & 65.11 & 65.11 & 60.64 \\
    Chapter-Llama & LLaMA-3.1-8B & ZS & 77.32 & 13.93 
    & 20.24 & 40.40 & 40.40 & 41.72 \\
    Chapter-Llama & LLaMA-3.1-8B & FT & 99.78 & 19.04 
    & 47.56 & 63.93 & 63.93 & 62.18 \\
    \midrule
    GPT-4o & -- & ZS & 1.82 & 2.34 
    & 24.32 & 24.89 & 26.01 & 35.23 \\
    GPT-4o-mini & -- & ZS & 2.80 & 2.95 
    & 15.88 & 21.50 & 34.27 & 31.21 \\
    Gemini-2.5-Pro & -- & ZS & 2.40 & 10.03 
    & 24.32 & 26.72 & 26.01 & 36.12 \\
    Gemini-2.0-Flash & -- & ZS & 0.28 & 3.07 
    & 9.64 & 11.71 & 27.99 & 13.54 \\
    Claude-3.5-Sonnet & -- & ZS & -- & 5.44 
    & 16.49 & 24.18 & 38.14 & 31.75 \\
    Claude-Sonnet-4 & -- & ZS & 0.02 & 10.93 
    & 18.08 & 25.36 & 34.35 & 38.48 \\
    \bottomrule
  \end{tabular}%
  }
  \caption{\label{tab:app_complete_baselines}
    Complete long-video LLM baseline and closed-source LLM reference results on AVLecture. ZS denotes zero-shot prompting and FT denotes fine-tuning. 
  }
\end{table*}

Table~\ref{tab:app_complete_baselines} provides the complete baseline results on AVLecture. Fine-tuned Chapter-Llama variants are much stronger than their zero-shot counterparts, indicating that task adaptation is important for long-video chaptering. Closed-source LLMs under zero-shot prompting achieve limited performance, especially on generation metrics, suggesting that simply prompting general-purpose LLMs is insufficient for this benchmark. These observations support the need for task-specific modeling of boundary localization and cross-segment context dependency.

\subsection{Closed-Source LLM Semantic Evaluation}
\label{app:closed_llm_semantic}

Lexical metrics can underestimate zero-shot closed-source LLMs when their outputs are semantically reasonable but use a different wording or granularity from the reference descriptions. We therefore complement CIDEr with a semantic-similarity evaluation. GPT-4o is given a generated chapter title and the corresponding reference title, and assigns a 0--100 semantic-similarity score based on topic match and specificity. Table~\ref{tab:closed_llm_semantic} reports the averaged scores. The semantic gap is smaller than the CIDEr gap, but fine-tuned chaptering models still achieve stronger semantic alignment with the AVLecture references.

\begin{table}[t]
  \centering
  \small
  \setlength{\tabcolsep}{6pt}
  \begin{tabular}{lc}
    \toprule
    \textbf{Method} & \textbf{Semantic similarity} \\
    \midrule
    CausalChapter & \textbf{82.10} \\
    Chapter-Llama & 79.80 \\
    Gemini-2.5-Pro & 60.40 \\
    \bottomrule
  \end{tabular}
  \caption{Semantic-similarity evaluation for chapter descriptions on AVLecture.}
  \label{tab:closed_llm_semantic}
\end{table}

\subsection{CSSE Output-Difference Sensitivity}
\label{app:csse_sensitivity}

The main method instantiates $D_{\mathrm{out}}$ as the teacher-forced token-level KL divergence between the full-context and leave-one-out predictive distributions. To assess whether CSSE depends on this particular distance function, we compare it with LLM-based relevance scoring, embedding distance, and token-overlap distance under the same Qwen2.5-3B setting. As shown in Table~\ref{tab:csse_sensitivity}, all variants improve over removing CSSE, while KL divergence performs best overall.

\begin{table}[t]
  \centering
  \small
  \setlength{\tabcolsep}{5pt}
  \begin{tabular}{lcc}
    \toprule
    \textbf{$D_{\mathrm{out}}$ implementation} & \textbf{CIDEr} & \textbf{SODA\_c} \\
    \midrule
    Without CSSE & 88.39 & 10.84 \\
    LLM-based scoring & 99.27 & 11.68 \\
    Embedding distance & 100.20 & 11.79 \\
    Token-overlap distance & 101.93 & 12.02 \\
    KL divergence & \textbf{104.59} & \textbf{12.73} \\
    \bottomrule
  \end{tabular}
  \caption{Sensitivity of CSSE to different output-difference functions on AVLecture.}
  \label{tab:csse_sensitivity}
\end{table}

\subsection{Inference Efficiency and Scalability}
\label{app:efficiency}

\paragraph{Profiling protocol and end-to-end comparison.}
We profile Chapter-Llama and CausalChapter on the same 10 AVLecture test videos using an NVIDIA A100 80GB GPU, BF16 precision, identical decoding settings, and the same warm-up procedure. The videos contain 4.3 predicted chapters on average. Chapter-Llama uses its original whole-video inference pipeline and jointly predicts boundaries and descriptions, so its two stages cannot be timed separately. CausalChapter first predicts boundaries and then applies intra-video micro-batching to chapter-level generation and leave-one-out scoring requests; the video-level batch size remains one. Table~\ref{tab:bf16_efficiency}(a) reports the matched end-to-end comparison using LLaMA-3.1-8B for both methods and a micro-batch size of 8 for CausalChapter.

\paragraph{Intra-video micro-batching.}
Once boundaries are fixed, requests from different chapters can be executed in parallel. We apply intra-video micro-batching to Pass 1 generation, full-context reference generation, leave-one-out KL scoring, and Pass 2 generation. Table~\ref{tab:bf16_efficiency}(b) reports the complete execution breakdown. Gen. and KL denote the average numbers of sequential batched model invocations per video after micro-batching; because they are averaged over videos, they need not be integers.

\begin{table}[!ht]
  \centering
  \small
  \textbf{(a) End-to-end comparison}\par\vspace{0.3em}
  \setlength{\tabcolsep}{3.2pt}
  \begin{tabular}{@{}lcccc@{}}
    \toprule
    \textbf{Method} & \textbf{Bound.} & \textbf{Gen.} & \textbf{Total} & \textbf{Mem.} \\
    \midrule
    Chapter-Llama & -- & 11.61 & 11.61 & 19.56 \\
    CausalChapter & 0.19 & 10.31 & 10.50 & 19.35 \\
    \bottomrule
  \end{tabular}

  \vspace{0.65em}
  \textbf{(b) Intra-video micro-batching}\par\vspace{0.3em}
  \setlength{\tabcolsep}{3.3pt}
  \begin{tabular}{@{}cccccc@{}}
    \toprule
    \textbf{Batch} & \textbf{Gen.} & \textbf{KL} & \textbf{CSSE} & \textbf{Total} & \textbf{Mem.} \\
    \midrule
    1 & 11.90 & 11.20 & 4.66 & 12.20 & 16.59 \\
    2 & 8.10 & 8.20 & 4.63 & 11.10 & 17.05 \\
    4 & 5.90 & 6.90 & 4.60 & 10.84 & 19.35 \\
    8 & 5.30 & 6.60 & 4.59 & 10.50 & 19.35 \\
    \bottomrule
  \end{tabular}
  \caption{BF16 inference on AVLecture. Times are seconds per video, memory is peak GiB, and Gen./KL are sequential batched calls per video. CausalChapter generation in (a) includes Pass 1, CSSE, and Pass 2.}
  \label{tab:bf16_efficiency}
\end{table}

At micro-batch size 1, the 11.90 generation calls in Table~\ref{tab:bf16_efficiency}(b) comprise 4.30 Pass 1 calls, 3.30 short CSSE reference-generation calls, and 4.30 Pass 2 calls; each reference-generation call produces at most 16 tokens. Increasing the micro-batch size to 8 reduces the sequential generation calls from 11.90 to 5.30 and KL calls from 11.20 to 6.60. CSSE time remains approximately 4.6 seconds per video and accounts for 4.59/10.50 (43.7\%) of the complete pipeline at size 8. Overall latency decreases from 12.20 to 10.50 seconds per video, while peak memory increases from 16.59 to 19.35 GiB.

\paragraph{Candidate-set scaling.}
Table~\ref{tab:csse_scaling}(a) evaluates the cost of expanding the pre-scoring candidate limit. The average number of available candidates saturates at 1.84 per chapter on the profiled videos. Consequently, increasing the limit from 1 to 10 changes the average number of KL calls only from 6.60 to 6.90 per video, while CSSE time increases from 4.43 to 5.31 seconds per video. This indicates that the practical intervention cost is governed by the compact set of available historical candidates rather than by the nominal limit alone.

\paragraph{Top-$K$ after intervention scoring.}
Top-$K$ is applied only after all candidates have been scored and therefore does not create additional intervention calls. Table~\ref{tab:csse_scaling}(b) reports its measured context count, time, and memory. Generation quality is reported once, in Figure~\ref{fig:topk_hyperparam} in the main paper.

\begin{table}[!ht]
  \centering
  \small
  \textbf{(a) Candidate-set limit}\par\vspace{0.3em}
  \setlength{\tabcolsep}{2.7pt}
  \begin{tabular}{@{}cccccc@{}}
    \toprule
    \textbf{Limit} & \textbf{Actual/ch.} & \textbf{KL} & \textbf{CSSE} & \textbf{Total} & \textbf{Mem.} \\
    \midrule
    1 & 0.77 & 6.60 & 4.43 & 9.93 & 19.37 \\
    3 & 1.60 & 6.60 & 5.24 & 10.82 & 19.36 \\
    5 & 1.81 & 6.90 & 5.51 & 10.96 & 19.36 \\
    10 & 1.84 & 6.90 & 5.31 & 10.71 & 19.56 \\
    \bottomrule
  \end{tabular}

  \vspace{0.65em}
  \textbf{(b) Top-$K$ after scoring}\par\vspace{0.3em}
  \setlength{\tabcolsep}{5pt}
  \begin{tabular}{@{}cccc@{}}
    \toprule
    \textbf{Top-$K$} & \textbf{Selected/ch.} & \textbf{Total} & \textbf{Mem.} \\
    \midrule
    1 & 0.77 & 10.71 & 19.56 \\
    3 & 1.60 & 10.98 & 19.50 \\
    5 & 1.81 & 10.84 & 19.35 \\
    7 & 1.84 & 10.79 & 19.38 \\
    9 & 1.84 & 10.43 & 19.58 \\
    \bottomrule
  \end{tabular}
  \caption{CSSE scaling on AVLecture. Times are seconds per video, memory is peak GiB, and KL denotes calls per video. In (a), Gen. calls remain 5.90 per video. Top-$K$ in (b) is applied after scoring and adds no leave-one-out evaluations.}
  \label{tab:csse_scaling}
\end{table}

End-to-end time remains within 10.43--10.98 seconds per video and peak memory within 19.35--19.58 GiB across the Top-$K$ sweep. As Figure~\ref{fig:topk_hyperparam} shows, $K=5$ gives the highest METEOR and a strong CIDEr score while using nearly all contexts available on average; we therefore use it as the balanced default.

\begin{figure}[!ht]
  \centering
  \includegraphics[width=\linewidth]{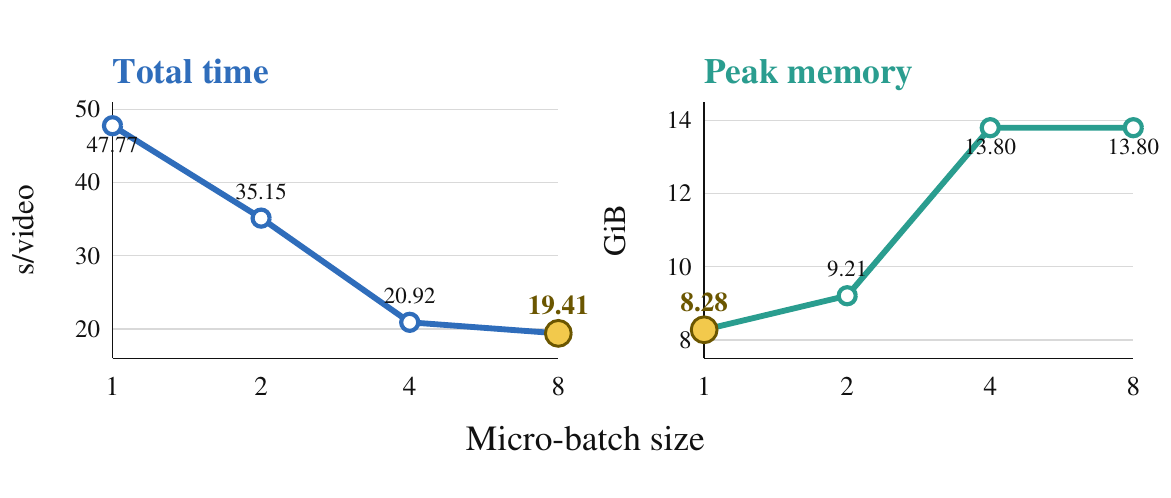}
  \caption{CausalChapter latency--memory trade-off under 4-bit NF4 quantization. Lower values are better; gold points mark the minimum in each panel.}
  \label{fig:microbatch_nf4}
\end{figure}

\paragraph{Quantized inference.}
We additionally profile 4-bit NF4 inference for memory-constrained settings. Figure~\ref{fig:microbatch_nf4} reports the latency--memory trade-off; each point is annotated with its exact value, so we do not repeat the measurements in a separate table.

At micro-batch size 1, quantized CausalChapter uses 8.28 GiB of peak memory, compared with 9.90 GiB for Chapter-Llama under the same NF4 setting. Increasing the size to 8 reduces latency from 47.77 to 19.41 seconds per video while using 13.80 GiB. These results provide an explicit latency--memory operating range rather than a single deployment point.

\FloatBarrier
\subsection{Qualitative Analysis}
\label{app:qualitative}
\begin{figure*}[t]
  \centering

  \begin{subfigure}{0.98\textwidth}
    \centering
    \includegraphics[width=\linewidth]{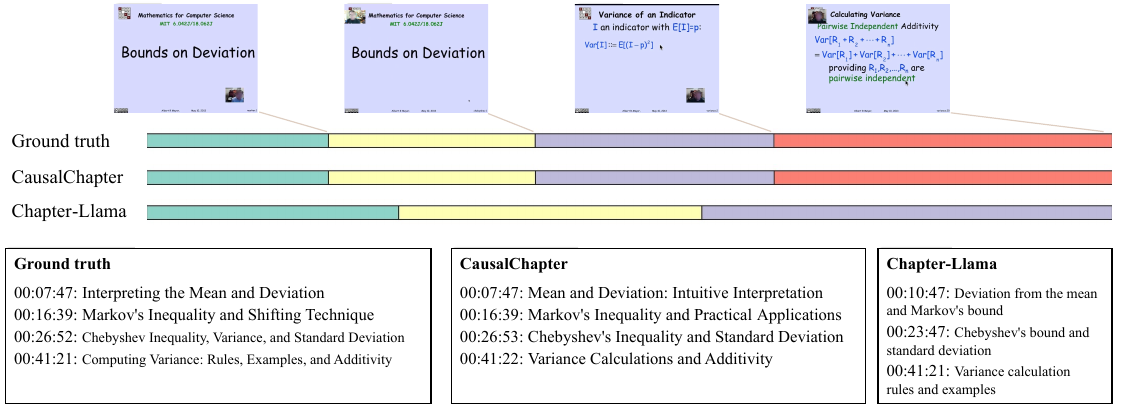}
    \caption{}
  \end{subfigure}

  \vspace{0.6em}

  \begin{subfigure}{0.98\textwidth}
    \centering
    \includegraphics[width=\linewidth]{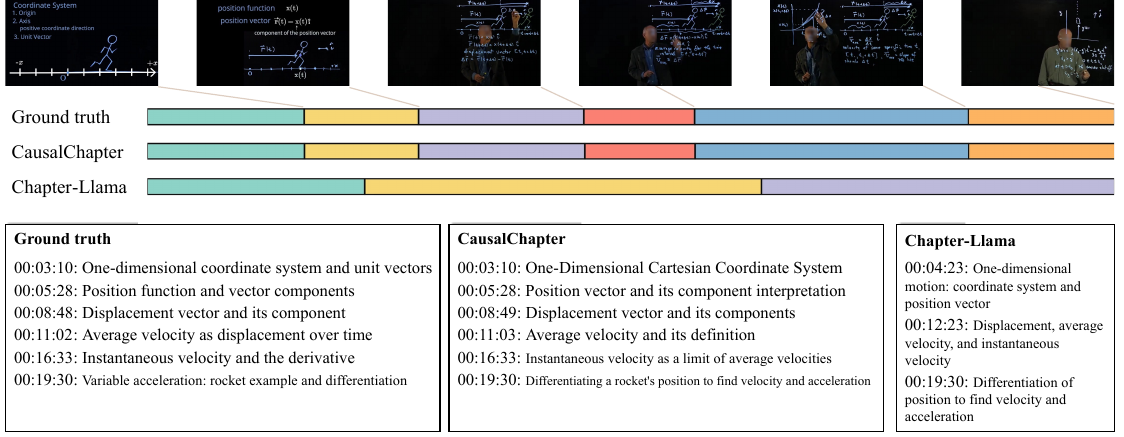}
    \caption{}
  \end{subfigure}

  \caption{
    Qualitative comparison on AVLecture. 
  }
  \label{fig:qualitative}
\end{figure*}
Beyond quantitative results, we further analyze representative cases in Figure~\ref{fig:qualitative} to examine how CausalChapter behaves under smooth boundary transitions and cross-segment context selection. 

In smooth-boundary cases, baseline methods often rely on local representation changes or textual similarity drops, and therefore tend to delay or miss boundaries when the topic gradually evolves. In contrast, CausalChapter captures a drop in intervention-defined predictive dependency between adjacent windows through LCDS, allowing it to identify structural changes more accurately. For chapter generation, similarity-based retrieval often selects contexts that are lexically close but largely repetitive. CSSE instead tends to select segments that provide definitions, background, experimental setup, or reasoning premises, leading to descriptions that are more complete and better aligned with the logical structure of the whole video.

The qualitative example is consistent with the quantitative results. LCDS provides a local dependency-shift signal for smooth boundary localization, while CSSE changes context selection from surface similarity matching to support estimation with respect to the current generation. These two behaviors help explain why CausalChapter improves both temporal localization and chapter-level generation.

\subsection{Prompt Design for Two-Pass Generation}

We adopt a two-pass prompting strategy for lecture subheading generation. In both
passes, the model receives the current segment transcript and generates one
concise subheading. The system instruction remains in the non-truncated prefix,
whereas the transcript occupies the truncatable middle. This preserves task
instructions, visual tokens, examples, contextual titles, and generation markers
under truncation. Table~\ref{tab:prompt_design} summarizes the two passes.

\FloatBarrier

\begin{table*}[t]
\centering
\small
\begin{tabular}{p{0.18\linewidth} p{0.76\linewidth}}
\toprule
\textbf{Component} & \textbf{Prompt Content} \\
\midrule
System Prompt
&
You are a helpful assistant generating lecture subheadings. Given the video
segment transcript, generate a concise and informative subheading that captures
the main topic. Only output the subheading, nothing else.
\\
\midrule
Pass 1:
Segment-Level Generation
&
Visual context: \texttt{<VIS\_0><VIS\_1>...<VIS\_N>}.

Optional few-shot examples are provided in the format:
\texttt{Segment: [example transcript] Subheading: [example title]}.

Optional related subheadings from the same lecture are provided as contextual
titles. The current segment transcript is then given as:
\texttt{Segment: [current segment transcript]}.

The model is required to output only the generated subheading.
\\
\midrule
Pass 2:
Temporal Refinement
&
Visual context: \texttt{<VIS\_0><VIS\_1>...<VIS\_N>}.

Temporal and causal rules: every retrieved context is from a segment before the
current segment. Use a retrieved context only if it helps clarify the topic
transition or dependency. Ignore irrelevant retrieved contexts. Never copy a
previous title as the current title. The final subheading must describe only the
current segment.

The Pass-1 generated title is provided as the initial draft. The model is asked
to keep it unchanged if it accurately captures the main concept using correct
technical terms. If the draft misses a critical technical term or the main
concept is wrong, the model generates a better subheading while preserving all
correct technical terms from the draft.

Retrieved previous contexts are provided as previous titles with metadata such
as temporal distance, source, and retrieval score. The current segment transcript
is then given as:
\texttt{Current segment transcript: [current segment transcript]}.

The model is required to output only the final refined subheading.
\\
\bottomrule
\end{tabular}
\caption{Prompt components used in the two-pass generation framework.}
\label{tab:prompt_design}
\end{table*}

\FloatBarrier

\section{LLM Usage Statement}
\label{llm}
We utilized a large language model (LLM) to improve the grammar, clarity, and overall readability of this manuscript. The LLM's role was strictly limited to language editing and polishing. All scientific contributions, including the core ideas, methodology, experimental design, data analysis, and conclusions, are the original work of the human authors. The use of the LLM did not alter the scientific content or its interpretation.

\end{document}